\documentclass[lettersize,journal]{IEEEtran}
\usepackage{amsmath,amsfonts}
\usepackage{algorithmic}
\usepackage{algorithm}
\usepackage{booktabs}
\usepackage{array}
\usepackage{textcomp}
\usepackage{stfloats}
\usepackage{url}
\usepackage{verbatim}
\usepackage{graphicx}
\usepackage{cite}
\usepackage{multirow}
\usepackage{multicol}
\usepackage{makecell}
\usepackage{subcaption}
\usepackage{bm}
\usepackage{tabularx}
\usepackage{color}
\begin{document}

\title{Adaptive Global-Local Representation Learning and Selection for Cross-Domain Facial Expression Recognition}

\author{Yuefang Gao, Yuhao Xie, Zeke Zexi Hu, Tianshui Chen, Liang Lin 
\thanks{This work was supported in part by National Key R\&D Program of China (Grant No.  2021ZD0111601), in part by the National Natural Science Foundation of China (Grant No. 62206060) and Guangzhou Basic and Applied Basic Research Foundation (Grant No. SL2022A04J01626). (Corresponding author: Tianshui Chen)}
\thanks{Yuefang Gao is with South China Agricultural University, Guangzhou, China. Yuhao Xie is with Guangzhou Institute of Technology, Xidian University, Guangzhou, China. Zeke Zexi Hu is with the School of Computer Science, University of Sydney, Darlington, New South Wales, Australia. Tianshui Chen is with Guangdong University of Technology, Guangzhou, China. Liang Lin is with Sun Yat-Sen University, Guangzhou, China.}}

\markboth{IEEE Transactions on Multimedia}%
{Shell \MakeLowercase{\tiextit{et al.}}: A Sample Article Using IEEEtran.cls for IEEE Journals}


\maketitle

\begin{abstract}
Domain shift poses a significant challenge in Cross-Domain Facial Expression Recognition (CD-FER) due to the distribution variation between the source and target domains. Current algorithms mainly focus on learning domain-invariant features through global feature adaptation, while neglecting the transferability of local features across different domains. Additionally, these algorithms lack discriminative supervision during training on target datasets, resulting in deteriorated feature representation in the target domain. To address these limitations, we propose an Adaptive Global-Local Representation Learning and Selection (AGLRLS) framework. The framework incorporates global-local adversarial adaptation and semantic-aware pseudo label generation to enhance the learning of domain-invariant and discriminative feature representation during training. Meanwhile, a global-local prediction consistency learning is introduced to improve classification results during inference. Specifically, the framework consists of separate global-local adversarial learning modules that learn domain-invariant global and local features independently. We also design a semantic-aware pseudo label generation module, which computes semantic labels based on global and local features. Moreover, a novel dynamic threshold strategy is employed to learn the optimal thresholds by leveraging independent prediction of global and local features, ensuring filtering out the unreliable pseudo labels while retaining reliable ones. These labels are utilized for model optimization through the adversarial learning process in an end-to-end manner. During inference, a global-local prediction consistency module is developed to automatically learn an optimal result from multiple predictions. To validate the effectiveness of our framework, we conduct comprehensive experiments and analysis based on a fair evaluation benchmark. The results demonstrate that the proposed framework outperforms the current competing methods by a substantial margin.
\end{abstract}


\section{Introduction}
Cross-Domain Facial Expression Recognition (CD-FER) aims at automatically determining a person’s emotional state from a face image, regardless of the domain, by transferring the learned knowledge from labeled source data to unlabeled target data. It plays a significant role in human-computer interaction \cite{lu2022domain}, affective computing \cite{9057390}, intelligent driving \cite{xia2020cross}, and so forth. While conventional facial expression recognition methods focus on a single dataset \cite{Wen2020,Huang2022,ShanLi2020DeepFE,ZhaoruiICPR2020}, CD-FER is more challenging due to the subtle interclass variations in different facial expression classes and the large discrepancy among FER datasets. These variations result in evident domain shift, which undermines models to generalize to more datasets. 

In the past decade, various CD-FER methods have been proposed and evaluated on popular FER datasets such as RAF-DB\cite{ShanLi2019ReliableCA}, FER2013\cite{IanGoodfellow2013ChallengesIR}, CK+ \cite{PatrickLucey2010TheEC}, JAFFE \cite{lyons1998coding}, SFEW2.0\cite{AbhinavDhall2011StaticFE}, and ExpW\cite{ZhanpengZhang2018FromFE}. These methods aimed to address the issue of data inconsistency between different datasets. Earlier studies have tackled this problem through transfer learning \cite{HaibinYan2016TransferSL} and supervised kernel matching \cite{miao2012cross}. However, these methods require a number of annotated samples from the target domain to ensure the discriminative information among categories, which is not suitable for unsupervised CD-FER settings. More recent strategies attempted to incorporate other learning techniques, including dictionary learning \cite{sun2022dynamic}, metric learning \cite{ni2020transfer}, and contrastive learning \cite{wang2022prototype}, to facilitate unsupervised CD-FER. Besides, some methods focused on generating more additional samples to bridge the gap in feature distribution between the source and target datasets \cite{XiaoqingWang2018UnsupervisedDA,YuanZong2018DomainRF}.

In recent years, significant efforts have been devoted to domain adaptation models for the CD-FER. These models aimed to learn transferable domain-invariant features by employing adversarial learning mechanisms \cite{bozorgtabar2020exprada,GuangLiang2020PoseawareAD,yang2018identity}. By aligning the feature distribution of the source and target domains, these methods achieved high recognition accuracy. However, despite their outstanding performance, these models primarily focused on extracting holistic features for domain adaptation, while overlooking the potential benefits of local features, e.g. greater transferability across domains and a more fine-grained representation of variations, which led to domain shift in inappropriate adversarial learning. To address the limitation, some studies \cite{YuanXie2020AdversarialGR,chen2021cross} combined graph representation propagation with adversarial learning to reduce the domain shift by modeling the correlation of holistic-local features within each domain and across different domains. There were also some works incorporating semantic information into the multi-view features learning to bridge the semantic gap during domain adaptation \cite{li2021jdman,xie2022learning}. These studies utilized the category prior knowledge or imposed the global and local semantic consistency constraints to learn semantically discriminative features. Nevertheless, the significant problem of imbalanced class distribution was largely overlooked among these methods, leading to suboptimal performance.

In this paper, we introduce an adaptive global-local representation learning and selection model to address the issue of data inconsistency, which adversely affects the recognition performance in cross-domain scenarios. The proposed method learns domain-invariant and discriminative feature representation in training by incorporating global-local adversarial adaptation and semantic-aware pseudo label generation, followed by a global-local prediction consistency strategy to refine recognition results in inference. Specifically, the model utilizes global and local adversarial learning independently to obtain compact and discriminative domain-invariant global and local features. Subsequently, a feature-level pseudo label generation mechanism is introduced to assign pseudo class labels to the features of unlabeled data, preventing mutual interference among classifiers. The separate classifier learning is further performed by optimizing the model with the adversarial learning process using these labels, facilitating the learning of more discriminative features for the target dataset. During inference, a global-local prediction consistency strategy is designed to learn the optimal class label by joint inference of the global and local prediction results. In this way, our approach significantly enhances the learned features, bolstering their generalization and representation capacities across the source and target datasets and meanwhile amplifying their discriminative power in the target domain, effectively mitigating domain shift to facilitate the CD-FER performance.

Building upon our previous conference work \cite{xie2022learning}, this paper presents several improvements. Firstly, instead of using a strict global-local semantic consistency constraint in the pseudo target label generation process, we adopt an adjustable threshold learning process to generate richer and more fine-grained pseudo labels for all the categories of the unlabeled faces, effectively alleviating the problem of class imbalance caused by source domain. Secondly, a global-local prediction consistency strategy is developed to infer the final labels of target data by combining the prediction scores from the global and local classifiers. This strategy mitigates the problem of domain shift across domains and achieves competitive performance. Finally, we conduct extensive experiments and ablation studies using more datasets and backbones to demonstrate the effectiveness of our method and the contribution of each component.

The contributions of this work can be summarized as follows: i) An Adaptive Global-Local Representation Learning and Selection (AGLRLS) model is proposed, which incorporates separate global-local adversarial learning and semantic-aware pseudo label generation to learn more domain-invariant and discriminative feature representation and thus address the domain shift and less discriminative ability issues in current CD-FER methods. ii) A semantic-aware pseudo label generation method is designed to produce reliable pseudo labels for each global and local feature of unlabeled data by utilizing learned adaptive thresholds. iii) A global-local prediction consistency learning is introduced that integrates global and local prediction results to infer the optimal class label, effectively bridging the semantic gap between source and target domains. iv) Comprehensive experiments are conducted to compare our proposed method with current CD-FER algorithms, demonstrating its superior performance. Codes and trained models are available at \textbf{\url{https://github.com/yao-papercodes/AGLRLS}}. 

\section{RELATED WORKS}
Quite a lot of research works have addressed the challenging task of cross-domain facial expression recognition. In this section, we mainly present a few methods that are closely related to this work, namely, cross-domain FER and adversarial domain adaptation. 

\subsection{Cross-Domain FER}
To deal with the data bias among different FER datasets, a series of CD-FER methods have been proposed \cite{HaibinYan2016TransferSL,Zou2022LearntoDecomposeCD,ni2020transfer,li2021jdman,XiaoqingWang2018UnsupervisedDA,YuanZong2018DomainRF,yan2016cross,Meng2023,zou2022facial,WenmingZheng2018CrossDomainCF,li2022deep}. Yan et al. \cite{HaibinYan2016TransferSL} utilized subspace learning to transfer the learned knowledge from the source data to the unlabeled target data. However, annotating some samples from the target data is necessary for this method. To handle CD-FER in unsupervised scenarios, work \cite{yan2016cross} presented a domain adaptive dictionary learning model that unified the unlabeled data to adaptively adjust the misaligned distribution in an embedded space. Later on, a transductive transfer regularized least-squares regression model was proposed \cite{WenmingZheng2018CrossDomainCF} that learned a discriminative subspace by combining the labeled source data and unlabeled auxiliary target data to reduce the dissimilarity of the marginal probability distribution between domains. In \cite{ni2020transfer}, a discriminative metric space was learned in a dictionary learning procedure to alleviate the influence of the distribution inconsistency. Different from these works, Wang et al. \cite{XiaoqingWang2018UnsupervisedDA} used samples generated by Generative Adversarial Network (GAN) on the target dataset to facilitate the CD-FER. Another domain regenerator was designed in \cite{YuanZong2018DomainRF} that regenerated source and target domain samples with the same or similar feature distribution to guide the label prediction of the unlabeled data. 

In the recently proposed methods, Ji et al. \cite{YanliJi2019CrossdomainFE} attempted to learn a common representation of expression in different domains through the fusion of intra-category common features and inter-category distinction features. \cite{ShanLi2020ADL} developed a deep emotion conditional adaptation network to learn the domain-invariant and class discriminative feature representation. This method aligned the marginal distribution globally and matched the fine-grained class-conditional distribution using the underlying label information on target datasets that can effectively mitigate the data bias between domains. In addition, Li et al. \cite{li2021jdman} proposed the deep margin-sensitive representation learning model that extracted semantically multi-level discriminative and transferable features by leveraging the category prior knowledge during domain adaptation to alleviate the domain shift. A recent study \cite{zou2022facial} introduced an emotion-guided similarity network that learned a transferable model for compound expression recognition from the unseen domain in the cross-domain few-shot learning scenario. In contrast to the current methods, we propose an adaptive global-local representation learning and selection method to address the problem of data inconsistency in CD-FER. 

\subsection{Adversarial Domain Adaptation}
Adversarial domain adaptation methods have recently become increasingly popular for cross-domain recognition due to the advantage of the disentangled and transferable representation learning 
\cite{bozorgtabar2020exprada,YuanXie2020AdversarialGR, chen2021cross,xie2022learning,EricTzeng2017AdversarialDD, conti20Cluster-level22cluster, de2022acdc,MingshengLong2018ConditionalAD,ji2021region,wang2021learning}. Inspired by the adversarial learning in generative adversarial networks, these methods utilize a feature extractor and a domain discriminator to mitigate domain shift through a two-player game in which the feature extractor learns the transferable domain-invariant features while the domain discriminator struggles to distinguish samples from the target dataset. As a pioneering effort, Tzeng et al. \cite{EricTzeng2017AdversarialDD} presented an adversarial discriminative domain adaptation method that combined discriminative modeling, united weight sharing, and adversarial loss to achieve better performance in challenging cross-modality classification task. Subsequent research \cite{MingshengLong2018ConditionalAD} applied the condition adversarial mechanism to the domain discriminator on the class information and proposed a conditional domain adversarial network with multilinear conditioning and entropy conditioning strategies to further improve the discriminability and transferability of the classifier. In \cite{GuangLiang2020PoseawareAD}, an unsupervised adversarial domain adaptation model was designed that integrated three learning strategies to adapt the pose and expression distribution between the source and target domain and learned the pose-and-identity features with robustness. Rather than focusing on holistic features for adaptation, some works \cite{YuanXie2020AdversarialGR,chen2021cross} integrated graph representation propagation \cite{li2015gated,chen2020knowledge} with adversarial learning to learn more representative and domain-invariant global-local features. Despite achieving superior performance on several publicly available face expression benchmark datasets, the learned features are less discriminative due to the lack of direct supervision for the samples of the target domain. 

Recent works show a tendency to incorporate the semantic-aware strategy while learning the feature representation\cite{bozorgtabar2020exprada,li2021jdman,xie2022learning,wang2023restoreformer++,pu2023spatial}. To reduce the semantic gap during domain distribution alignment, Bozorgtabar et al. \cite{bozorgtabar2020exprada} employed adversarial domain adaptation to transform the visual appearance of the images from different domains while preserving the semantic information. In \cite{xie2022learning}, a consistent global-local and semantic learning method was proposed that integrated the domain-invariant global-local features and consistent semantic learning to further mitigate the problem of semantic inconsistency during domain adaptation. However, the method employed fixed criteria in the pseudo label generation process, which could lead to only a limited number of expression classes having the capability to generate pseudo labels.  In contrast to the previously mentioned methods, our proposed approach focuses on the domain-invariant multi-scale features through separate global and local adversarial learning and preserves the underlying semantic consistency by the global-local unified prediction selection strategy.

\section{Proposed method}
\begin{figure*}[htbp]
  \centering
  \includegraphics[width=0.95\linewidth,height=0.60\linewidth]{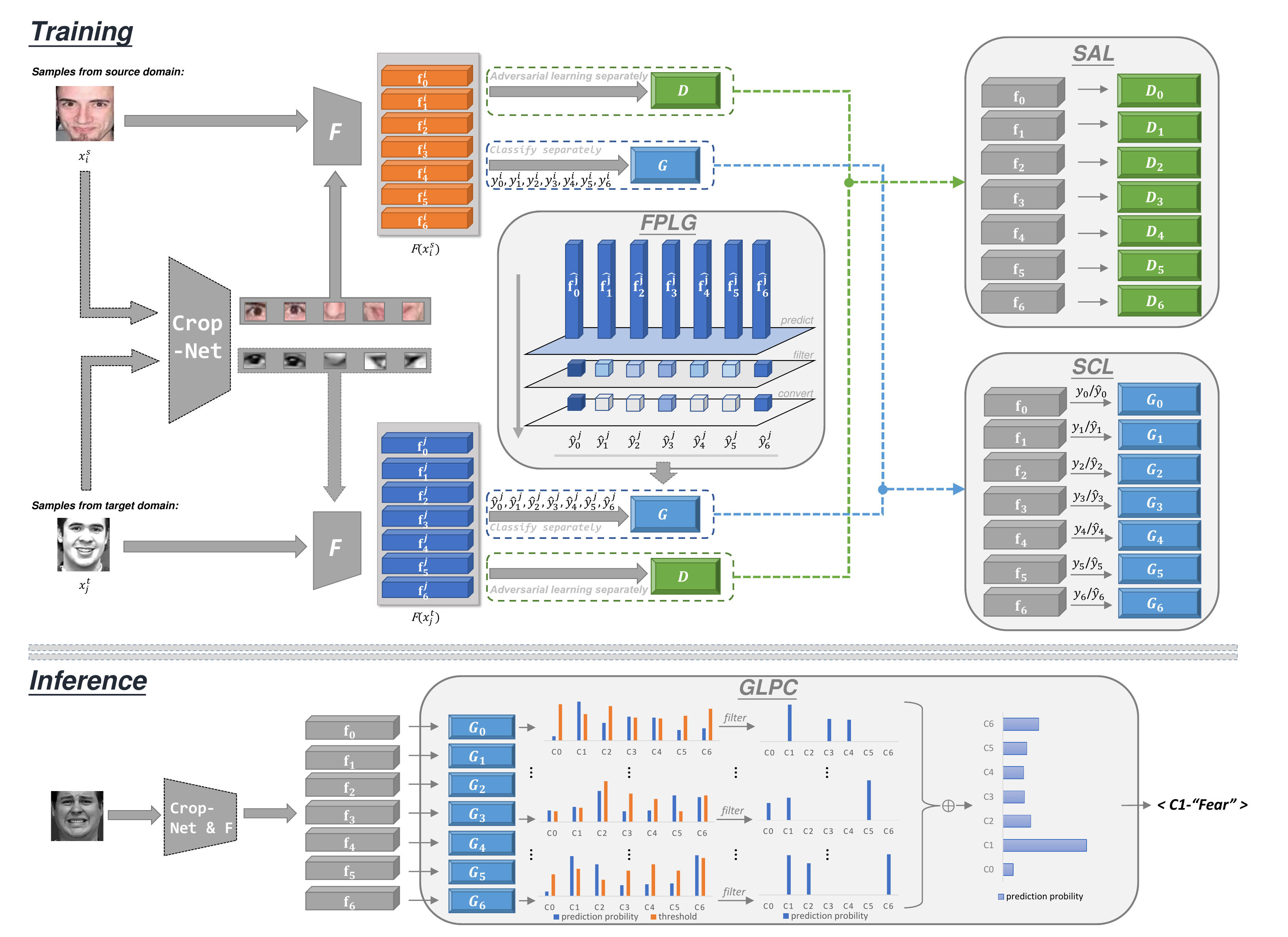}

  \caption{Illustration of the training and inference stages of our proposed AGLRLS model.}
  \label{fig:framework}
\end{figure*}
\subsection{Overview}
Our objective is to tackle the Cross-Domain Facial Expression Recognition (CD-FER) task where we are given a source dataset $\mathcal{D}_{s}=\left\{\left(x_{i}^{s}, y_{i}^{s}\right)\right\}_{i=1}^{n_{s}}$ and a target dataset $\mathcal{D}_{t}=\left\{\left(x_{j}^{t}\right)\right\}_{j=1}^{n_{t}}$. Here, $n_s$ and $n_t$ represent the sample quantities in the source and target domains, respectively. Each sample $x_i^s$ from the source dataset, which is used for training, is associated with a label $y_i^s$. The samples from the target domain have no labels. To this end, we propose an adaptive global-local feature representation learning and selection framework that can effectively mitigate the data inconsistency and bias through the following two-player games as
\begin{equation}\label{eq:MinMax-Discriminator}
    \begin{split}
    \min_{D} \mathcal{L}\left(F, D\right),
    \end{split}
\end{equation}
\begin{equation}\label{eq:MinMax-Classifier&Extractor}
\begin{split}
    \min _{G, F} \mathcal{L}(F, G)-\mathcal{L}\left(F,D\right)
\end{split}
\end{equation}
where
\begin{equation}\label{eq:Train-Discriminator-Detailed}
    \begin{split}
        \mathcal{L}\left(F,D\right)=\sum_{i=0}^{k}&-\mathbb{E}_{x^s\sim\mathcal{D}_{s}}\log{\left[D_i\left(F\left(x^{s}\right)\right)\right]}\\
        &-\mathbb{E}_{x^t \sim \mathcal{D}_{t}}\log{\left[1-D_i\left(F\left(x^{t}\right)\right)\right]},
    \end{split}
\end{equation}
\begin{equation}\label{eq:Train-Classifier&Extractor-Detailed}
\begin{split}
    \mathcal{L}\left(F, G\right)=\sum_{i=0}^{k}&-\mathbb{E}_{\left(x^{s}, y^{s}\right) \sim \mathcal{D}_{s}} \ell\left(\textit{G}_i\left(F\left(x^{s}\right)\right), y^{s}\right)\\
    &-\mathbb{E}_{\left(x^t, \hat{y^t}\right) \sim \mathcal{D}_{t}} \ell\left(\textit{G}_i\left(F\left(x^{t}\right)\right), \hat{y^{t}}\right)
\end{split}
\end{equation}
Here, $\textit{F}$ represents the feature extractor, while $G_i$ and $D_i$ belong to the sets of classifiers $G$ and domain discriminators $D$, respectively. The variable $k$ denotes the total number of extracted features, which is equivalent to the number of members in $G$ and $D$. $\ell$ represents the classification loss function, and $\mathbb{E}$ stands for the expectation operation. $\hat{y^{t}}$ represents the pseudo labels of the target samples. Equations (\ref{eq:MinMax-Discriminator}) and (\ref{eq:MinMax-Classifier&Extractor}) describe the objective of the feature extractor, which is to extract ambiguous features from input images in order to confuse the domain discriminators. Conversely, the domain discriminator aims to discern the domain to which the input features belong. Through repeated iterations of the adversarial learning process, the feature extractor gradually reduces domain bias and extracts domain-invariant features for cross-domain recognition. Specifically, for each input image $\textit{x}$, we partition it into global and local regions, which are then input to $\textit{F}$ to generate corresponding global and local features. These features are subsequently fed into their respective domain discriminator and classifier to facilitate domain-invariant representation learning. To enhance the feature extractor's ability to extract discriminative domain-invariant features, we introduce a feature-level pseudo label generation module. This module can generate abundant and reliable semantic labels based on individual global and local features from the target data. As a result, the incorporation of these pseudo labels into the training process further enhances the generalization performance of classifiers $G$. During inference, we propose a global-local prediction consistency module that automatically selects the optimal prediction from multiple generated labels. Fig. \ref{fig:framework} provides a visual illustration of the aforementioned procedures, and additional details will be elaborated in the subsequent sections.

\subsection{Domain-Invariant Representation Learning}
\subsubsection{Separate Adversarial Learning}\label{SAL}
In this subsection, we will present a detailed explanation of the separate adversarial learning process, denoted as SAL in Fig.  \ref{fig:framework}. As the previous work \cite{chen2021cross} has suggested, local regions surrounding specific facial landmarks play a crucial role in expression recognition. Building upon this insight, we adopt a strategy to extract not only the full-face image as the global image but also the key facial landmarks as the local images. These landmarks include the left eye (le), right eye (re), nose (ne), left mouth corner (lm), and right mouth corner (rm). Following the aforementioned procedure, we obtain a global feature vector, denoted as $\mathbf{f_g}$, as well as five local feature vectors, namely $\mathbf{f_{le}}$, $\mathbf{f_{re}}$, $\mathbf{f_{ne}}$, $\mathbf{f_{lm}}$, and $\mathbf{f_{rm}}$. Additionally, we concatenate these six vectors into a seventh vector, denoted as $\mathbf{f_{gl}}$. Consequently, a feature set $\textbf{f}$ is formed, comprising seven vectors. The strategy can be mathematically expressed as follows,
\begin{equation}\label{eq:construct-features}
    \begin{split}
    &\mathbf {f_{gl}}=\mathbf{f_g}\oplus \mathbf{{f_{le}}}\oplus \mathbf{{f_{re}}}\oplus \mathbf{{f_{ne}}}\oplus \mathbf{{f_{lm}}}\oplus \mathbf{{f_{rm}}} ,\\
    &\mathbf{f}=\left \{\mathbf{f_g}, \mathbf{{f_{le}}},\mathbf{{f_{re}}},\mathbf{{f_{ne}}},\mathbf{{f_{lm}}},\mathbf{{f_{rm}}}, \mathbf{f_{gl}} \right \}
    \end{split}
\end{equation}
where $\oplus$ is the concatenation operation. To perform separate adversarial learning, we utilize indices ranging from 0 to 6 to access the vector $\textbf{f}$. Thus, Equation (\ref{eq:Train-Discriminator-Detailed}) can be formulated as below,
\begin{equation}\label{eq:Train-Discriminator-Detailed2}
    \begin{split}
        \mathcal{L}\left(F,D\right)=\sum_{i=0}^{6}&-{\beta_i \cdot} \mathbb{E}_{\mathbf{f_i^s} \sim\mathcal{D}_{s}}\log{\left[D_i\left(\mathbf{f_i^s}\right)\right]}\\
        &-{\beta_i \cdot} \mathbb{E}_{\mathbf{f_i^t} \sim \mathcal{D}_{t}}\log{\left[1-D_i\left(\mathbf{f_i^t}\right)\right]}
    \end{split}
\end{equation}
{$\mathbf{f_i^s}$ represents a member feature vector in the feature set $\mathbf{f^s}$ which is generated by the feature extractor $F$ when processing an image from the source domain.} Likewise, the feature vector $\mathbf{f_i^t}$ corresponds to an image from the target domain. Specifically, the expectation loss is computed based on the entropy at the feature level. By applying binary classification cross-entropy using $\log{\left[D_i\left(\mathbf{f_i}\right)\right]}$, we obtain a measure of the loss. {$\beta_i$ is a coefficient within the balance coefficient set $\beta$ used to balance the loss values based on $\mathbf{f_i^s}$ and $\mathbf{f_i^t}$}. Notably, our proposed method leverages both global and local features within the input $\mathbf{f}$, facilitating the learning of domain-invariant features at both the global and local levels. As a result, the method significantly reduces domain bias.

\subsubsection{Feature-level Pseudo Label Generation}\label{FPLG}
Current works \cite{chen2021cross,YuanXie2020AdversarialGR} utilize domain-invariant features derived from an adversarial learning process for classification, directly applying these features without incorporating target dataset supervision. This approach, relying solely on source dataset labels for classifier training, potentially diminishes the discriminative capacity, which can result in suboptimal performance. In this work, we present an innovative pseudo label generation algorithm that generates reliable pseudo labels for samples in the target dataset, substantially boosting the discriminative performance in the target dataset.

Instead of using the traditional image-level generation approach, where each image is assigned a pseudo label, we employ a feature-level paradigm that produces more comprehensive semantic labels for the unlabeled data in the target domain. To achieve this, we introduce the Feature-level Pseudo Label Generation (FPLG) module in Fig. \ref{fig:framework}, which independently generates a corresponding pseudo label for each learned domain-invariant feature. Specifically, in Equation (\ref{eq:construct-features}), seven features are obtained from a target data $x^t$, each of which is subsequently fed into the corresponding classifier $G_i$ to make a prediction,
\begin{equation}\label{eq:Prediction-Score}
    s^i = G_i(\mathbf{f_i^t})=\left\{s^i_j \right\}, {\text{s.t.} j=0,..,c-1.}
\end{equation}
where {$s^i$ represents the total prediction score of $G_i$ based on $\mathbf{f_i^t}$ and $s^i_j$ is the score that $G_i$ predicts $x^t$ as class $j$ based on the $\mathbf{f_i^t}$. $c$ indicates the number of categories.} The predicted class $p$ is obtained from $s^i$ with the maximum probability,
\begin{equation}
    p=argmax(s^i)
\end{equation}
Then, with the prediction result of each classifier, we need to determine whether the generated pseudo label is valid. Inspired by FlexMatch \cite{zhang2021flexmatch}, a dynamic threshold set $t^i$ is applied to each classifier,
\begin{equation}\label{eq:Threshold}
    t^i = \left \{ t^i_j \right \}, \text{s.t.} j=0,..,c-1.
\end{equation}
where $t^i_j$ represents the threshold that $G_i$ needs to reach to generate pseudo label for class $j$. $t^i_j$ is calculated by the following equation,
\begin{equation}\label{eq:ThresholdCalculate}
    t^i_j = \mathcal{M}(\lambda^i_j) \times \theta
\end{equation}
where $\theta$ is a fixed threshold, $\mathcal{M}$ is a nonlinear mapping function with respect to $\lambda^i_j$, and $\lambda^i_j$ represents the proportion of data that the classifier $G_i$ has successfully generated pseudo label for class $j$, which can be calculated as follows,

\begin{equation}\label{eq:CollectNum}
    \lambda^i_j=\frac{\sigma^i_j}{\max \sigma^i}
\end{equation}
where $\sigma^i$ represents the cumulative pseudo label quantities generated by $G_i$ for all categories until current training process, while $\sigma^i_j$  signifies the accumulation for class $j$. To ensure the sensitivity of $\mathcal{M}$ to $\lambda^i_j$, we improve the $\mathcal{M}$ function so that the mapped value is always greater than $\lambda^i_j$, 
\begin{equation}\label{eq:map-function}
    \mathcal{M}(\lambda^i_j) = \frac{(\lambda^i_j+1)^2}{4}
\end{equation}
Finally, we can obtain the final value of $\hat{y_i}$ by comparing $s^i_{p}$ and $t^i_{p}$ through the following formula,
\begin{equation}\label{eq:pl-value}
    \hat{y_i}=\left\{\begin{array}{ll}
        p ,& s_{p}^{i} > t_{p}^{i} \\
        -1  ,& otherwise
\end{array}\right.
\end{equation}
where -1 indicates that the generation of a pseudo label has failed. We repeat the above process seven times and obtain seven pseudo labels $\hat{y}$ based on the extracted seven features.
\begin{equation}
    \hat{y} = \left \{ \hat{y_0}, \hat{y_1}, \hat{y_2}, \hat{y_3}, \hat{y_4}, \hat{y_5}, \hat{y_6} \right \}
\end{equation}

Upon the reliable pseudo labels flexibly taking the class imbalance issue into consideration and containing abundant semantic information, we leverage these target data to retrain the network with the SCL module. This joint training process not only boosts the discriminability of features but also enhances the generalization of the model across domains. Specifically, Equation (\ref{eq:Train-Classifier&Extractor-Detailed}) captures the essence of this training process, which can be further refined as shown below,
\begin{equation}\label{eq:Train-Classifier&Extractor-Detailed2}
\begin{split}
    \mathcal{L}(F, G)=\sum_{i=0}^{6}&-\eta_i \cdot \mathbb{E}_{\left(\mathbf{f_i^s}, y^{s}_{i}\right) \sim \mathcal{D}_{s}} \ell\left(\textit{G}_i\left(\mathbf{f_i^s}\right), y^{s}_{i}\right)\\
    &-\eta_i \cdot \mathbb{E}_{\left(\mathbf{f_i^t}, \hat{y^t_i}\right) \sim \mathcal{D}_{t}} \ell\left(\textit{G}_i\left(\mathbf{f_i^t}\right), \hat{y^t_i}\right)
\end{split}
\end{equation}
$y^s_i$ denotes the label of $\mathbf{f^s_i}$ from the source domain while $\hat{y^t_i}$ denotes the one from the target domain as described in Equation (\ref{eq:pl-value}). It is worth noting that while $y^s_i$ shares the same value within a sample, $\hat{y^t_i}$ might vary as the pseudo label generation is feature-level and classifiers may exhibit disagreement. We use the expectation of entropy as the classification loss to train the feature extractors and classifiers. $\eta_i$ is a coefficient within the balance coefficient set $\eta$ used to equalize the loss values associated with $\mathbf{f^s_i}$ and $\mathbf{f^t_i}$.

\subsection{Global-Local Prediction Selection Learning}\label{Global-local predicting consistently}
In inference, to make an accurate and unified prediction of the target data, a Global-Local Prediction Consistency (GLPC) module is proposed to aggregate the predictions of the seven classifiers while taking their consistency into consideration. As shown in the GLPC module in Fig. \ref{fig:framework}, for each target domain image $x^t$, each classifier obtains a set of prediction scores $s^i$, as mentioned in Equation (\ref{eq:Prediction-Score}). 
Similar to the FPLG module in the training process, a set of threshold scores $t^i$ is calculated for each classifier as in Equation (\ref{eq:Threshold}). As a result, a 7$\times$7 score matrix $\mathbf{s}$ and a 7$\times$7 threshold matrix $\mathbf{t}$ are obtained as
\begin{equation}\label{eq:Scores&Threshold}
\begin{split}
    \mathbf{s}=\left(\begin{array}{cccc}
    s^0_0 & s^0_1 & \cdots & s^0_6 \\
    s^1_0 & s^1_1 & \cdots & s^1_6 \\
    \vdots & \vdots & \ddots & \vdots \\
    s^6_0 & s^6_1 & \cdots & s^6_6
    \end{array}\right),
    \mathbf{t}=\left(\begin{array}{cccc}
    t^0_0 & t^0_1 & \cdots & t^0_6 \\
    t^1_0 & t^1_1 & \cdots & t^1_6 \\
    \vdots & \vdots & \ddots & \vdots \\
    t^6_0 & t^6_1 & \cdots & t^6_6
    \end{array}\right)
\end{split}
\end{equation}
where we assume there are 7 categories of facial expression. As the target domain images with pseudo labels in the training set share a similar distribution to the data in the testing set, the knowledge learned from FPLG can be transferred to the inference stage as prediction consistency. The final predicted class $p$ is initially determined by selecting the class with the highest prediction score from the classifier $G_6$, which takes global-local features $\mathbf{f_{gl}}$ as input. Formally,
\begin{equation}\label{eq:pseudolabels-gl}
\begin{split}
p = argmax(s^6)
\end{split}
\end{equation}
However, this selection process, as described in Equation (\ref{eq:pseudolabels-gl}), only applies if the score $s^6_p$ is greater than the threshold $t^6_p$. Otherwise, the prediction class of $s^0$ from the classifier $G_0$, which utilizes global features $\mathbf{f_g}$, will be considered. Similarly, the predicted class $p$ from $G_0$ will serve as the final prediction only if the score $s^0_p$ is greater than the threshold $t^0_p$. If neither of these two classifiers provides a confident prediction, the final prediction class will be determined by aggregating the five local predictions with the above two predictions ($s^0$ and $s^6$). Specifically, we calculate a mask matrix $\mathbf{m}$ using the following formula,
\begin{equation}
\begin{split}
\mathbf{m} = \mathbf{s} > \mathbf{t} =
\left(\begin{array}{cccc}
    1 & 0 & \cdots & 1 \\
    1 & 0 & \cdots & 1 \\
    \vdots & \vdots & \ddots & \vdots \\
    0 & 1 & \cdots & 0
    \end{array}\right)
\end{split}
\end{equation}
The matrix $\mathbf{m}$, which is also a 7$\times$7 matrix, consists of elements $\mathbf{m_{ij}}$. $\mathbf{m_{ij}}$ takes a value of 1 if $s^i_j$ is greater than $t^i_j$, otherwise, it is set to 0. Next, the masked score matrix $\mathbf{\hat{s}}$ is obtained by the dot product of $\mathbf{s}$ and $\mathbf{m}$.
\begin{equation}
\begin{split}
\mathbf{\hat{s}} = \mathbf{s} \cdot \mathbf{m} =
\left(\begin{array}{cccc}
    \hat{s^0_0} & 0 & \cdots & \hat{s^0_6} \\
    \hat{s^1_0} & 0 & \cdots & \hat{s^1_6} \\
    \vdots & \vdots & \ddots & \vdots \\
    0 & \hat{s^6_1} & \cdots & 0
    \end{array}\right)
\end{split}
\end{equation}
The prediction score for each class is produced by summing up the masked scores from all the classifiers as follows,
\begin{equation}
\begin{split}
\mathbf{\tilde{s}} = \left \{ \sum\limits_{i=0}^6 \mathbf{\hat{s^i_j}} \right \}, j=0...6.
\end{split}
\end{equation}
Finally, our predicted label $p$ is the class with the highest prediction score in $\mathbf{\Tilde{s}}$ as
\begin{equation}
\begin{split}
p = argmax(\mathbf{\Tilde{s}})
\end{split}
\end{equation}

\section{Experiments}

\subsection{Experimental Settings}
\subsubsection{Dataset}
\begin{figure}[t!]
  \centering
  \includegraphics[width=0.8\linewidth]{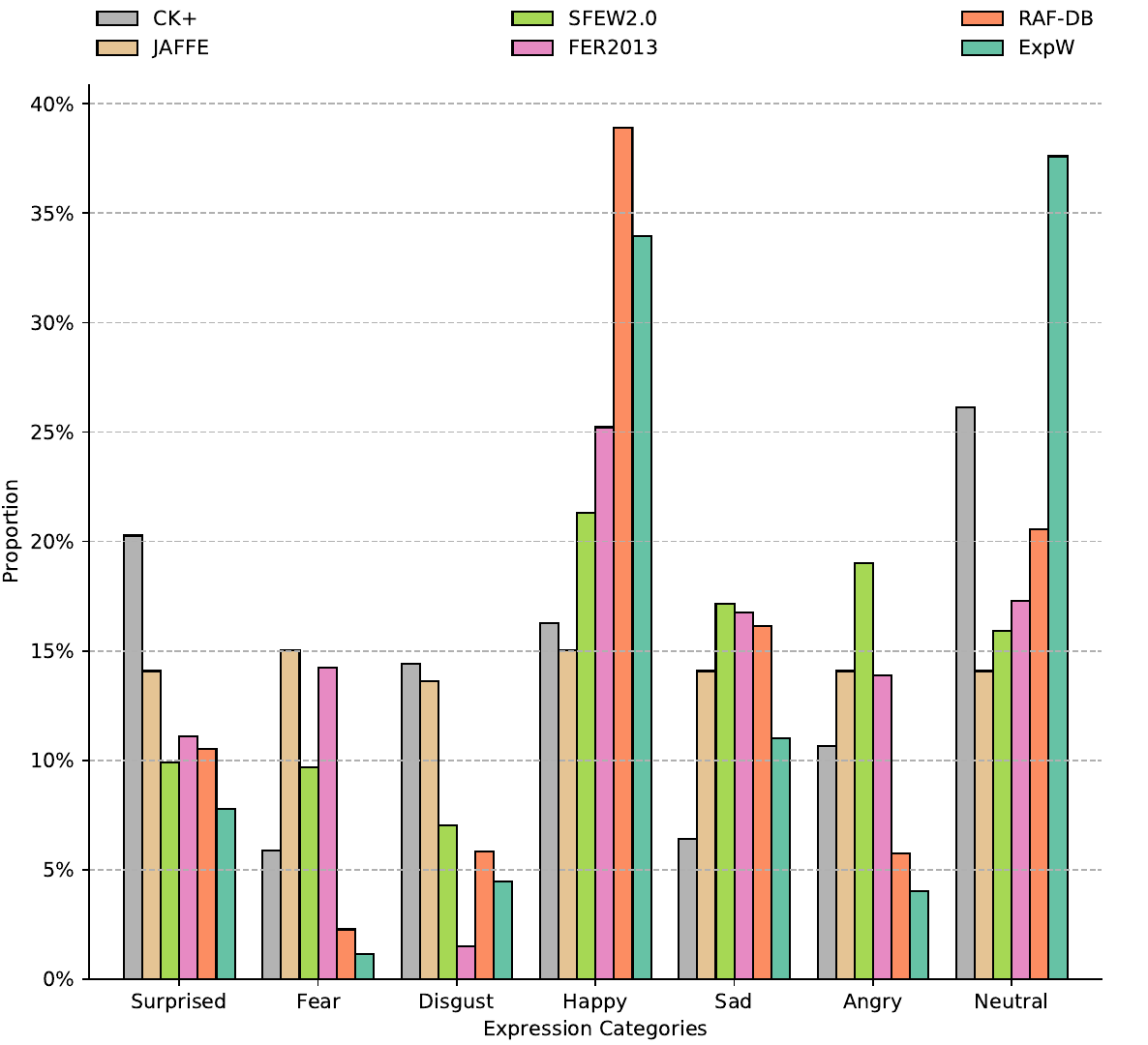}
  \caption{Bar plot of category distribution in the training datasets.}
  \label{fig:distribution}
\end{figure}

There are six datasets used in the experiments, namely RAF-DB\cite{ShanLi2019ReliableCA}, FER2013\cite{IanGoodfellow2013ChallengesIR}, CK+\cite{PatrickLucey2010TheEC}, JAFFE\cite{lyons1998coding}, SFEW2.0\cite{AbhinavDhall2011StaticFE}, and ExpW\cite{ZhanpengZhang2018FromFE}. $\textbf{RAF-DB}$ includes 29,672 diverse, gender-balanced facial images across various ages and poses. $\textbf{FER2013}$, compiled via the Google image search engine, contains 35,887 images of different facial expressions. $\textbf{CK+}$ offers 593 annotated video samples from 123 subjects, used for lab-based facial expression recognition. $\textbf{JAFFE}$ features 213 images from 10 Japanese women, each of which is labeled as one of six basic expressions or as neutral. $\textbf{SFEW2.0}$ is an 'in-the-wild' dataset with unconstrained facial expressions, varying in age, pose, and resolution. $\textbf{ExpW}$ has 91,793 'in-the-wild' face images, collected via Google's search API. The class quantity distribution of the aforementioned datasets is illustrated in Fig. \ref{fig:distribution}.

\subsubsection{Evaluation Metrics}\label{evaluation-metric}
To conduct fair comparison experiments, multiple evaluation metrics are used in this paper  as follows:

\textbf{Accuracy} represents the percentage of samples that are correctly predicted. It is also known as overall accuracy and is the most commonly used metric in the classification task.

\textbf{Recall} is the proportion of true positive prediction among all actual positive instances. Although it was originally used for binary classification task, it can be also applied to multi-classification task and effectively measure a model's performance, even in a class-imbalanced dataset. In our case, we utilize Macro-Average Recall to obtain the average recall of all categories as the final result. For simplicity, we will refer to Macro-Average Recall as Recall hereafter.

\textbf{Precision} is the proportion of true positive prediction among all positive predictions made by a model. It can serve as a metric to evaluate a model's performance in a class-imbalanced dataset as well. Similar to  Recall, we use Macro-Average Precision by averaging the values for classes to obtain the final Precision.

\textbf{F1 Score} is a metric that strikes a balance between Recall and Precision in the classification task. Similar to Recall and Precision, we adopt the Macro-Average F1 Score in our approach, which involves calculating the F1 score for each class using Equation (\ref{eq:f1}) and then averaging these values to obtain the final F1 Score.
\begin{equation}\label{eq:f1}
\begin{split}
\textit{F1\,score} = \frac{2 \times recall \times precision}{recall + precision}
\end{split}
\end{equation}

\textbf{Mean Metric} refers to the average value of various metrics across datasets, including Mean Accuracy, Mean Recall, Mean Precision, and Mean F1 Score. These results provide a comprehensive measurement of the algorithm's performance encompassing a diverse range of datasets.

\begin{table*}
  \centering
    \caption{Accuracies of our proposed AGLRLS and existing leading methods on various datasets using different source datasets and backbone networks. The best results are in bold.}
  \label{tab:CMP-SOTA-OURS}
  \begin{tabular}{c|cccccc|cccccc}
  \toprule
  \centering \multirow{2}*{Method} & \multicolumn{6}{c}{i) Source=RAF-DB, Backbone=ResNet50} & \multicolumn{6}{|c}{ii) Source=RAF-DB, Backbone=MobileNet-v2}\\
  \cline{2-13}
  \centering ~ & CK+ & JAFFE & SFEW2.0 & FER2013 & ExpW & Mean & CK+ & JAFFE & SFEW2.0 & FER2013 & ExpW & Mean\\
  \hline
  \centering  DT \cite{chen2021cross} & 71.32 & 50.23 & 50.46 & 54.49 & 67.45 & 58.79 & 66.67 & 38.97 & 41.74 & 49.99 & 63.08 & 52.09\\
  \centering  PLFT \cite{chen2021cross} & 77.52 & 53.99 & 48.62 & 56.46 & 69.81 & 61.28 & 72.09 & 38.97 & 41.97 & 51.11 & 64.12 & 53.65 \\
  \centering  ICID \cite{YanliJi2019CrossdomainFE} & 74.42 & 50.70 & 48.85 & 53.70 & 69.54 & 59.44 & 57.36 & 37.56 & 38.30 & 44.47 & 60.64 & 47.67\\
  \centering  DFA \cite{RonghangZhu2016DiscriminativeFA} & 64.26 & 44.44 & 43.07 & 45.79 & 56.86 & 50.88 &41.86 & 35.21 & 29.36 & 42.36 & 43.66 & 38.49\\
  \centering  FTDNN \cite{MarcusViniciusZavarez2017CrossDatabaseFE} & 79.07 & 52.11 & 47.48 & 55.98 & 67.72 & 60.47 & 71.32 & 46.01 & 45.41 & 49.96 & 62.87 & 55.11\\
  \centering  JUMBOT \cite{pmlr-v139-fatras21a} & 79.46 & 54.13 & 51.97 & 53.56 & 63.69 & 60.56 & 73.64 & 51.35 & 44.41 & 49.05 & 60.84 & 55.86\\
  \centering  ETD \cite{MengxueLi2020EnhancedTD} & 75.16 & 51.19 & 52.77 & 50.41 & 67.82 & 59.47 & 69.27 & 48.57 & 41.34 & 49.43 & 57.05 & 53.13\\
  \centering  CADA \cite{MingshengLong2018ConditionalAD} & 72.09 & 52.11 & 53.44 & 57.61 & 63.15 & 59.68 & 62.79 & 53.05 & 43.12 & 49.34 & 59.40 & 53.54\\
  \centering  SAFN \cite{RuijiaXu2019LargerNM} & 75.97 & 61.03 & 52.98 & 55.64 & 64.91 & 62.11 & 66.67 & 45.07 & 40.14 & 49.90 & 61.40 & 52.64\\
  \centering  SWD \cite{ChenYuLee2019SlicedWD} & 75.19 & 54.93 & 52.06 & 55.84 & 68.35 & 61.27 & 68.22 & 55.40 & 43.58 & 50.30 & 60.04 & 55.51\\
  \centering  LPL \cite{li2017reliable} & 74.42 & 53.05 & 48.85 & 55.89 & 66.90 & 59.82 & 59.69 & 40.38 & 40.14 & 50.13 & 62.26 & 50.52\\
  \centering  DETN \cite{ShanLi2018DeepET} & 78.22 & 55.89 & 49.40 & 52.29 & 47.58 & 56.68 & 53.49 & 40.38 & 35.09 & 45.88 & 45.26 & 44.02\\
  \centering  ECAN \cite{ShanLi2020ADL} & 79.77 & 57.28 & 52.29 & 56.46 & 47.37 & 58.63 & 53.49 & 43.08 & 35.09 & 45.77 & 45.09 & 44.50\\
  \centering  AGRA \cite{chen2021cross} & 85.27 & 61.50 & 56.43 & 58.95 & 68.50 & 66.13 & 72.87 & 55.40 & 45.64 & 51.05 & 63.94 & 57.78\\
  \centering  CGLRL \cite{xie2022learning} & 82.95 & 59.62 & 56.88 & 59.30 & 70.02 & 65.75 & 69.77 & 52.58 & 49.77 & 52.46 & 64.87 & 57.89\\
  \hline
  \centering  AGLRLS & \textbf{87.60} & \textbf{61.97} & \textbf{58.26} & \textbf{60.68} & \textbf{73.00} & \textbf{68.30} & \textbf{82.95} & \textbf{56.81} & \textbf{50.23} & \textbf{54.51} & \textbf{69.10} & \textbf{62.72}\\
  \bottomrule

\toprule
  \centering \multirow{2}*{Method} & \multicolumn{6}{c}{iii) Source=FER2013, Backbone=ResNet50} & \multicolumn{6}{|c}{iv) Source=FER2013, Backbone=MobileNet-v2}\\
  \cline{2-13}
  \centering ~ & CK+ & JAFFE & SFEW2.0 & RAF-DB & ExpW & Mean & CK+ & JAFFE & SFEW2.0 & RAF-DB & ExpW & Mean\\
  \hline
  \centering  DT \cite{chen2021cross} & 68.99 & 44.13 & 42.43 & 63.84 & 54.17 & 54.71 & 62.02 & 39.44 & 30.96 & 40.95 & 47.05 & 44.08 \\
  \centering  PLFT \cite{chen2021cross} & 79.07 & 46.48 & 42.20 & 67.92 & 54.96 & 58.13 & 61.24 & 44.60 & 28.67 & 40.30 & 53.63 & 45.69 \\
  \centering  ICID \cite{YanliJi2019CrossdomainFE} & 63.57 & 44.60 & 43.58 & 62.08 & 54.01 & 53.57 & 55.81 & 39.44 & 31.42 & 41.21 & 41.50 & 41.88\\
  \centering  DFA \cite{RonghangZhu2016DiscriminativeFA} & 55.81 & 42.25 & 34.86 & 48.84 & 44.55 & 45.26 & 55.81 & 36.15 & 27.78 & 34.18 & 43.83 & 39.55 \\
  \centering  FTDNN \cite{MarcusViniciusZavarez2017CrossDatabaseFE} & 72.09 & 53.99 & 45.64 & 64.40 & 54.67 & 58.16 & 59.69 & 45.54 & 39.68 & 52.43 & 49.87 & 49.44 \\
  \centering  JUMBOT \cite{pmlr-v139-fatras21a} & 75.76 & 49.69 & 44.33 & 63.08 & 52.37 & 57.05 & 51.16 & 41.54 & 36.06 & 44.93 & 49.30 & 44.60 \\
  \centering  ETD \cite{MengxueLi2020EnhancedTD} & 77.42 & 44.17 & 39.58 & 61.18 & 49.97 & 54.46 & 54.55 & 40.32 & 30.77 & 50.54 & 45.91 & 44.42 \\
  \centering  CADA \cite{MingshengLong2018ConditionalAD} & 81.40 & 45.07 & 46.33 & 65.96 & 54.84 & 58.72 & 66.67 & 50.23 & 41.28 & 53.15 & 51.84 & 52.63 \\
  \centering  SAFN \cite{RuijiaXu2019LargerNM} & 68.99 & 45.07 & 38.07 & 62.80 & 53.91 & 53.77 & 66.67 & 37.56 & 35.78 & 38.73 & 45.56 & 44.86 \\
  \centering  SWD \cite{ChenYuLee2019SlicedWD} & 65.89 & 49.30 & 45.64 & 65.28 & 56.05 & 56.43 & 53.49 & 48.36 & 35.78 & 47.44 & 50.02 & 47.02 \\
  \centering  LPL \cite{li2017reliable} & 68.22 & 42.72 & 44.27 & 64.23 & 52.45 & 54.38 & 60.47 & 37.56 & 31.88 & 43.92 & 49.83 & 44.73 \\
  \centering  DETN \cite{ShanLi2018DeepET} & 65.89 & 37.89 & 37.39 & 50.51 & 52.15 & 48.77 & 48.09 & 42.31 & 27.54 & 40.53 & 39.14 & 39.52 \\
  \centering  ECAN \cite{ShanLi2020ADL} & 60.47 & 41.76 & 46.01 & 53.41 & 48.88 & 50.11 & 55.65 & 44.12 & 28.46 & 42.31 & 41.53 & 42.41 \\
  \centering  AGRA \cite{chen2021cross} & 85.69 & 52.74 & 49.31 & 67.62 & 60.23 & 63.12 & 67.44 & 47.89 & 41.74 & 52.27 & \textbf{59.41} & 53.75 \\
  \centering  CGLRL \cite{xie2022learning} & 79.84 & 53.52 & 52.29 & 71.84 & 61.94 & 63.87 & 59.69 & 50.23 & 44.72 & 61.95 & 55.33 & 54.38\\
  \hline
  \centering  AGLRLS & \textbf{89.92} & \textbf{54.93} & \textbf{52.52} & \textbf{72.02} & \textbf{62.63} & \textbf{66.40} & \textbf{69.77} & \textbf{50.70} & \textbf{47.02} & \textbf{64.85} & 57.14 & \textbf{57.90}\\
  \bottomrule
  \end{tabular}
  \vspace{4pt}
\end{table*}

\subsubsection{Implementation}
Here, we will provide the implementation details of our framework.
\paragraph{Network Architecture}
In our approach, we utilize two backbone models, ResNet50-variant \cite{KaimingHe2016DeepRL} and MobileNet-v2 \cite{DBLP:conf/cvpr/SandlerHZZC18}, as feature extractors, following the method proposed in \cite{chen2021cross}. Both models consist of four block layers. When provided with an input image of size 112$\times$112, the second layer of the network generates feature maps $\textit{m}_1$ of size 28$\times$28$\times$128, while the fourth layer produces feature maps $\textit{m}_2$ of size 7$\times$7$\times$512. For extracting global features, we apply a convolution operation to transform $\textit{m}_2$ into feature maps of size 7$\times$7$\times$64. Subsequently, an average pooling layer is employed to obtain a feature vector of 64 dimensions. Regarding local features, we employ MT-CNN \cite{KaipengZhang2016JointFD} to crop five regions of size 7$\times$7$\times$128 from $\textit{m}_1$, centered around the corresponding facial landmarks. Similar to global features, these local feature maps are processed with a convolutional operation and average pooling to generate five local feature vectors of 64 dimensions. To combine the global and local features, we concatenate the global feature vector with the five local feature vectors, resulting in a feature vector of 384 dimensions. The set of feature vectors, denoted as $\textbf{f}$, consists of seven elements. Corresponding to the seven features in $\textbf{f}$, seven classifiers $G$ and domain discriminators $D$ are constructed, the former of which uses a series of fully connected layers and the latter uses the approach in \cite{MingshengLong2018ConditionalAD}.

\paragraph{Training Details}
In the proposed AGLRLS framework, the feature extractors $F$, classifiers $G$ and domain discriminators $D$ are trained and optimized in Equations (\ref{eq:MinMax-Discriminator}) and (\ref{eq:MinMax-Classifier&Extractor}). We initialize the backbone network using models pretrained on the MS-Celeb-1M dataset \cite{YandongGuo2016MSCeleb1MAD}. The parameters of the newly added layers are initialized using the Xavier algorithm \cite{XavierGlorot2010UnderstandingTD}. Stochastic Gradient Descent (SGD) is used as the optimizer in the experiment. Inspired by the previous works \cite{YuanXie2020AdversarialGR, JunWen2019ExploitingLF}, we adopt a two-stage training process. In the first stage, we solely utilize labeled source domain samples to train the feature extractor and classifier using cross-entropy loss, as described in the first part of Equation (\ref{eq:Train-Classifier&Extractor-Detailed2}). The learning rate, weight decay, and momentum are set to 0.0001, 0.0005, and 0.9, respectively. We train the network for approximately 15 epochs. In the second stage, the domain discriminator is trained utilizing the objective loss specified in Equation (\ref{eq:Train-Discriminator-Detailed2}), while the feature extractor and classifier undergo fine-tuning using the objective loss described in Equation (\ref{eq:Train-Classifier&Extractor-Detailed2}). The momentum and weight decay parameters remain unchanged from the first stage. Specifically, the initial learning rate for the feature extractor and the source classifier is set to 0.00001 and subsequently divided by 10 after 20 epochs. As for the domain discriminator, it is trained from scratch with an initial learning rate of 0.0001, and this learning rate is reduced by a factor of 10 once the error reaches a saturation point. The balance coefficient set $\beta$ and $\eta$ are both set to $\left \{7,1,1,1,1,1,7  \right \}$.

In addition, in the pseudo label generation module, three points should be noted. First, although the pseudo label generation and cross-entropy calculation share the same target domain images, data augmentation is performed differently as the former uses weak augmentation and the latter applies stronger enhancement such as diffraction. Second, after using the classifier to predict the initial label scores, we apply a softmax process to compare with the threshold. Thirdly, pseudo label generation is seamlessly integrated into the model's forward propagation, constituting a minor portion of the total runtime. In our experiments, the average forward propagation time for a single batch is 0.78 seconds, while the generation of seven sets of pseudo labels and the associated calculations require only 0.05 seconds, accounting for approximately 6.41\%.

\subsection{Comparisons With The Existing Methods}\label{Comparisons with the existing methods}
In the context of cross-domain facial expression recognition, comparing the performance of different methods can be challenging due to variations in the source domain and backbones. To ensure a fair comparison, we refer to the benchmark proposed by Chen et al. \cite{chen2021cross} and compare our proposed AGLRLS method with a number of leading methods. The comparative results are presented in Table \ref{tab:CMP-SOTA-OURS}. It is evident from the table that our method consistently outperforms the other algorithms across various configurations and datasets. For more insights, we will delve into a comprehensive analysis of our proposed method's performance under different conditions.

\subsubsection{Using Different Backbones}
The utilization of different backbones results in variations in the capabilities of our feature extractor. To assess the robustness of our method with different backbones, we employed two widely used backbones, namely ResNet50 and MobileNet-v2. The comparative results presented in Table \ref{tab:CMP-SOTA-OURS} demonstrate that when we keep the source domain constant and replace ResNet50 with MobileNet-v2, the performances of almost all the approaches show a varying degree of degradation across all datasets. For a more detailed analysis, we focus on subtables i and ii in Table \ref{tab:CMP-SOTA-OURS}, where RAF-DB is the source domain and MobileNet-v2 replaces ResNet50. The AGRA algorithm, previously regarded as the best, experiences a reduction of 12.4\%, 6.1\%, 10.79\%, 7.9\%, and 4.56\% in accuracy on the CK+, JAFFE, SFEW2.0, FER2013, and ExpW datasets, respectively. Similarly, our preliminary work, CGLRL, also shows decreases of 13.18\%, 7.04\%, 7.11\%, 6.84\%, and 5.15\% on these five datasets, respectively. This reduction in performance can be attributed to the fact that MobileNet-v2 utilizes depthwise separable convolution in its architecture, resulting in relatively weaker feature extraction capability compared to ResNet50. While the performance of our proposed AGLRLS method also declines, the reduction is comparatively smaller than that of AGRA and CGLRL. Specifically, we observe decreases of 4.65\%, 5.16\%, 8.03\%, 6.17\%, and 3.90\% on the aforementioned datasets, respectively. Notably, even when using MobileNet-v2 with relatively inferior feature extraction capability, our method still outperforms the best algorithm on the five datasets by 9.31\%, 1.41\%, 0.46\%, 2.05\%, and 4.23\%, respectively. To facilitate a fair comparison, we calculate the average accuracy across each target domain, referred to as the Mean. This demonstrates the consistent robustness of AGLRLS across the target datasets.

\subsubsection{Using Different Source Domain}
Another important factor that affects the accuracy of the target domain is the similarity between the source and target domain. As shown in Table \ref{tab:CMP-SOTA-OURS}, when we maintain the same backbone network and switch the source domain from RAF-DB to FER2013, almost all of the algorithms experience varying degrees of performance degradation across all datasets. For example, in subtables i) and iii) of Table \ref{tab:CMP-SOTA-OURS}, we use the same backbone, ResNet50, and change the source domain from RAF-DB to FER2013.
The accuracy of AGRA increases by 0.42\% on CK+ dataset but decreases by 8.76\%, 7.12\%, and 8.27\% on JAFFE, SFEW2.0, and ExpW datasets, respectively. Similarly, CGLRL experiences reductions of 3.11\%, 6.1\%, 4.59\%, and 8.08\%. The reasons can be summarized as follows: 

Firstly, although RAF-DB and FER2013 have similar data sizes and are both collected from the Web, FER2013 data are in grayscale while RAF-DB data are in color. The target datasets, SFEW2.0 and ExpW, consist of color images. From this perspective, the similarity between FER2013 and SFEW2.0/ExpW is smaller compared to the similarity between RAF-DB and SFEW2.0/ExpW. Similarly, CK+ is also a grayscale image dataset, and its similarity with FER2013 is higher than the similarity between CK+ and RAF-DB. Consequently, some methods may show a slight improvement in performance on FER2013. The second difference between RAF-DB and FER2013 is the different coverage of ethnic groups. The RAF-DB dataset includes a considerable number of samples of Asian people, while FER2013 does not. As a result, the similarity between FER2013 and the JAFFE dataset, which consists of Asian women, is much lower compared to the similarity between RAF-DB and JAFFE. This explains the significant performance drop of most methods on the JAFFE dataset.

Despite experiencing some degree of degradation, our proposed model still outperforms other state-of-the-art approaches on CK+, JAFFE, SFEW2.0, and ExpW by 4.23\%, 1.41\%, 0.23\%, and 0.69\%, respectively. To ensure a fair comparison, we utilize the Mean metric for evaluation. As shown in Table \ref{tab:CMP-SOTA-OURS} (iii), our AGLRLS method consistently achieves the best performance compared to the other approaches.

\subsubsection{Using Statistical Test}\label{Using Mean accuracy and statistical test}

\begin{figure}[t!]
  \centering
  \includegraphics[width=1.0\linewidth]{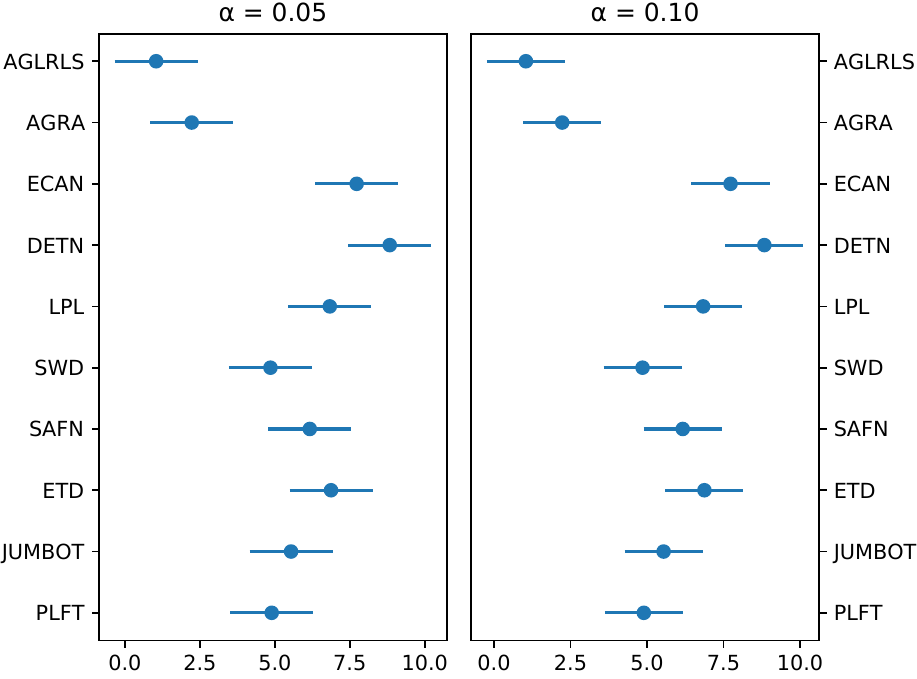}
  \caption{The Friedman Test Chart. The left half is the case of $\alpha$=0.05, and the right half is the case of $\alpha$=0.10. The horizontal coordinate corresponding to the intermediate point of each algorithm is the average order value. The lower the value, the better the performance. The range of horizontal lines on both sides of each intermediate point represents the CD value. If the horizontal lines between the two algorithms do not overlap, it means that the performance of these two methods is significantly different.}
  \label{fig:nemenyi}
\end{figure}

Statistical tests offer a more rigorous and scientific approach for comparing algorithm performance, enabling the assessment of significant differences rather than relying solely on direct accuracy comparisons. In our study, we employ two well-established statistical methods, namely the Friedman test \cite{Friedman1937} and the Nemenyi test \cite{Tamura1968Dist}, to demonstrate the superiority of our algorithm. However, due to the limited number of accuracy results available for each algorithm in each backbone and source domain configuration, the small dataset size may result in a lack of statistical significance. To address this issue, we aggregate the accuracies and Mean values of each method across different backbone and source domain configurations, resulting in a collection of 24 accuracy data points for each algorithm. From the set of algorithms listed in Table \ref{tab:CMP-SOTA-OURS}, we select 10 algorithms with competitive performance for comparative testing. We calculate the average rank of each method following the approach described in \cite{Friedman1937} and then employ the Nemenyi test as a post-hoc analysis. To facilitate the interpretation of the results, we compute the critical difference (CD) using the following formula,
\begin{equation}\label{eq:CD_Calu}
    \begin{split}
    CD=q_{\alpha} \sqrt{\frac{k(k+1)}{6 n}}
    \end{split}
\end{equation}
Here, the $\alpha$ represents the confidence level, which is a parameter in statistics that is manually set and commonly takes values such as 0.05 and 0.10. The parameter $n$ corresponds to the number of samples available for each algorithm, while $k$ represents the total number of algorithms being compared. The value of $q_{\alpha}$ can be obtained by referring to the work \cite{Demsar2006StatisticalComp}, which provides critical values for various significance levels.

The Friedman test chart in Fig. \ref{fig:nemenyi} displays the average rank of algorithms and the CD. Upon inspection, it is evident that our method achieves the highest average rank among all the approaches. Notably, there is no overlap between our AGLRLS and most of the other methods except for AGRA. This signifies that our method, AGLRLS, exhibits a statistically significant difference from the other approaches. However, some overlap is observed between AGLRLS and AGRA since AGRA ranks closely behind AGLRLS by only a 1-rank difference. We attribute this result to the limited number of accuracy data points (24 in total). Nonetheless, when calculating the CD values using Equation (\ref{eq:CD_Calu}), we obtain the values of 2.77 and 2.55 which are both greater than 1. 

\subsubsection{Using Class Imbalance Metrics}

\begin{figure*}[htbp]
  \centering
  \includegraphics[width=1.0\linewidth]{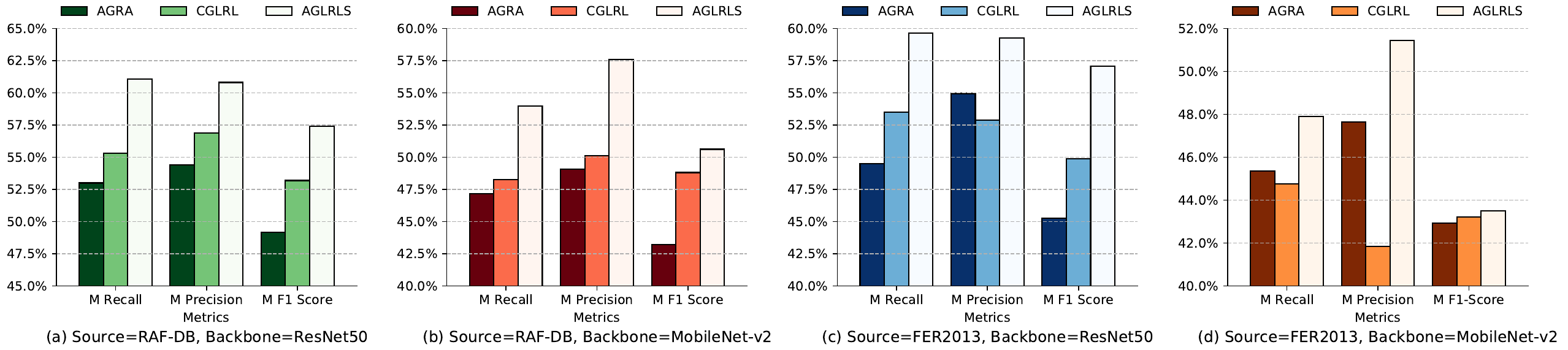}
  \caption{ Bar charts of the three methods' performance under the four configurations.}
  \label{fig:3metrics}
\end{figure*}
As shown in Fig. \ref{fig:distribution}, the facial expression dataset utilized in this study exhibits a class imbalance. Therefore, evaluating the algorithm solely based on overall accuracy metrics may be susceptible to influence from high-frequency categories, potentially resulting in an inadequate reflection of the model's performance. Consequently, we employ class imbalance metrics, including Recall, Precision, and F1 Score, to comprehensively evaluate the model's robustness across all categories. Taking into account the necessity to evaluate the effectiveness in the five target domains for each backbone network and source domain configuration, we use Mean Recall (M Recall), Mean Precision (M Precision), and Mean F1 Score (M F1 Score) as metrics for a  more intuitive comparison. As depicted in Fig. \ref{fig:3metrics}, our method demonstrates leading performance compared to AGRA and CGLRL on all configurations and metrics. This implies that our proposed algorithm exhibits enhanced recognition robustness for each expression category, showcasing its potential to adapt to more complex facial expression recognition tasks, especially in real-world applications.

\subsection{Ablation Study}\label{AblativeStudy}
In this subsection, we will conduct a series of experiments to analyze the contribution of each module in our proposed  AGLRLS model, and then focus on a more detailed analysis of our feature-level pseudo label generation module and global-local prediction consistency module. Subsequent experiments utilize the ResNet50 backbone, with RAF-DB as the source domain, evaluating on CK+, JAFFE, SFEW2.0, FER2013, and ExpW datasets.

\subsubsection{Contribution of Modules in AGLRLS}
\begin{table}
  \centering
    \caption{Accuracies on five target domains after adding each module incrementally. S, F, and G represent the SAL, FPLG, and GLPC modules, respectively. The best results are in bold.}
    \label{tab:Module-Contribution}
  \begin{tabular}{c@{\hspace{-2em}}cc|ccccc|c}
  \toprule
  \centering  S & F & G & CK+ & JAFFE & SFEW2.0 & FER2013 & ExpW & Mean\\
  \hline
  \centering  $\times$ & $\times$ & $\times$ & 73.64 & 59.15 & 52.29 & 56.88 & 68.93 & 62.18\\
  \centering  $\checkmark$ &  $\times$ &  $\times$ & 75.19 & 59.62 & 52.52 & 57.92 & 70.22 & 63.90\\
  \centering  $\checkmark$ & $\checkmark$ &  $\times$ & 86.82 & 61.03 & 57.34 & 60.51 & 72.96 & 67.73\\
  \centering  $\checkmark$ & $\checkmark$ & $\checkmark$ & \textbf{87.60} & \textbf{61.97} & \textbf{58.26} & \textbf{60.68} & \textbf{73.00} & \textbf{68.30}\\
  \bottomrule
  \end{tabular}
  \vspace{4pt}
\end{table}

Our proposed AGLRLS model mainly has three novel modules: Separate Adversarial Learning (SAL), Feature-level Pseudo Label Generation (FPLG) and Global-Local Prediction Consistency (GLPC). To establish a foundation for the analysis, we set up a baseline by performing separate classification learning using only the labeled source domain data. The accuracy for the CK+, JAFFE, SFEW2.0, FER2013, and ExpW datasets are 73.64\%, 59.15\%, 52.29\%, 56.88\%, and 68.93\%, respectively, as indicated in the initial row of Table \ref{tab:Module-Contribution}. Based on this, we add the SAL module to learn domain-invariant features. As a result, the accuracy improves by 1.55\%, 0.47\%, 0.23\%, 1.04\%, and 1.29\%, respectively. These gains underscore the efficacy of domain adaptive adversarial learning in aligning feature spaces between the source and target domains. Building upon this, we incorporate the FPLG module to exploit target domain data. The resultant accuracies, presented in the third row of Table \ref{tab:Module-Contribution}, show substantial improvements: 11.63\%, 1.41\%, 4.82\%, 2.59\%, and 2.74\% across the five datasets. This significant improvement can be attributed to the substantial number of reliable pseudo labels generated by the FPLG module. Integrating these labels into the training process markedly enhances the discriminative capacity of domain-invariant representation, facilitating the classification performance in the target domain. Lastly, we introduce the GLPC module, as illustrated in the fourth row of Table \ref{tab:Module-Contribution}. This addition leads to incremental accuracy gains of 0.78\%, 0.94\%, 0.92\%, 0.17\%, and 0.04\%. We attribute these gains to GLPC's ability to harmonize predictions from seven classifiers with their consistency at both global and local levels.

\subsubsection{Comparisons of Different Pseudo Label Generation Strategies}
\begin{figure*}[htbp]
  \centering
  \includegraphics[width=1.0\linewidth]{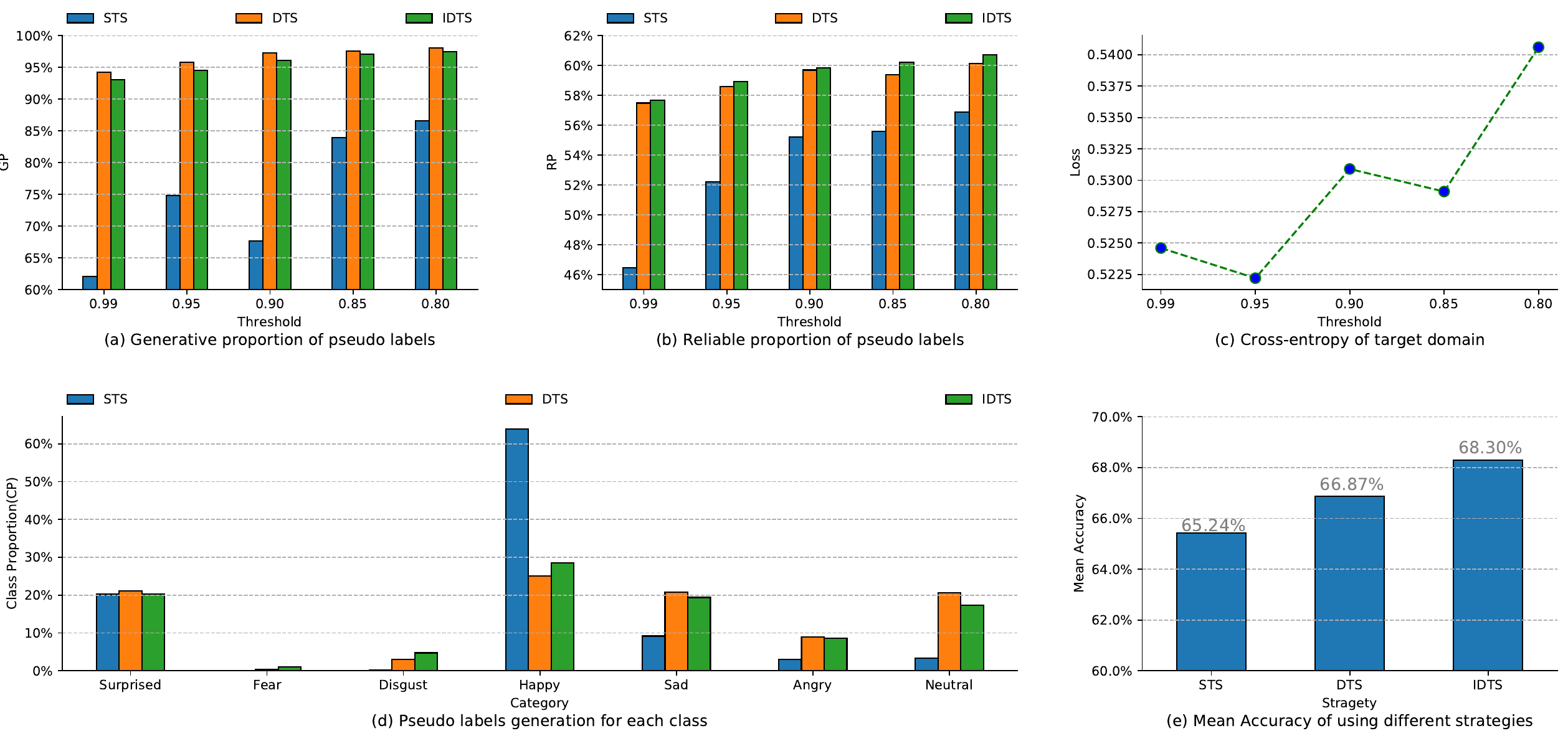}
  \caption{Ablation analysis of FPLG. (a) and (b) show the percentage of the number of pseudo labels and reliable pseudo labels that can be generated by three different pseudo label generation strategies. (c) shows that the pseudo labels generated by IDTS are used to calculate the cross-entropy loss of the target domain data. (d) shows the percentage of pseudo labels of each category generated by the three strategies for each category, relative to the total number of generated pseudo labels. (e) shows the Mean accuracy of the FPLG module using the three strategies.}
  \label{fig:fplg-analysis}
\end{figure*}

From Table \ref{tab:Module-Contribution}, we can observe that the FPLG module has the most significant contribution to AGLRLS. The foundation of the FPLG module lies in the adoption of the Improved Dynamic Threshold Strategy (IDTS). We systematically compare methods to understand their performance factors. Based on domain-invariant features extracted by the SAL module, we apply three strategies for generating pseudo labels: Static Threshold Strategy (STS), Dynamic Threshold Strategy (DTS) from \cite{zhang2021flexmatch}, and Improved Dynamic Threshold Strategy (IDTS) from subsection \ref{FPLG}. IDTS improves upon DTS by applying a novel non-linear mapping function to effectively filter low-quality pseudo labels during early training.

We use three comparison metrics of pseudo labels: Generative Proportion (GP), Reliable Proportion (RP), and Category Proportion (CP). Evaluation is based on average values across five target domains. Firstly, in Fig. \ref{fig:fplg-analysis}(a), DTS and IDTS achieve higher GP than STS across thresholds due to dynamic threshold adjustments based on pseudo label proportions, particularly for challenging classes, leading to increased pseudo label generation in the target domain. Moreover, Fig. \ref{fig:fplg-analysis}(b) shows RP variations under distinct thresholds, highlighting dynamic threshold strategy's superior reliability, particularly with IDTS, benefiting from an improved mapping function in Equation (\ref{eq:map-function}) improving pseudo label accuracy. Furthermore, in Fig. \ref{fig:fplg-analysis}(b), RP rises as thresholds decrease due to increasing GP (Fig. \ref{fig:fplg-analysis}(a)). To mitigate wrong pseudo labels' impact, cross-entropy loss guides a fixed threshold. Fig. \ref{fig:fplg-analysis}(c) shows IDTS-generated pseudo labels' cross-entropy loss at varying thresholds, with the lowest at 0.95, adopted as default. Additionally, Fig. \ref{fig:fplg-analysis}(d) shows class-wise pseudo label generation via three strategies. STS correlates with samples; Happy (ample data) achieves CP $>$ 60\%, while Fear and Disgust (scarce samples) barely generate. DTS and IDTS yield more balanced distributions, offering reliability, regardless of samples, implying potential for model enhancement via pseudo labels. Lastly, in Fig. \ref{fig:fplg-analysis}(e), Mean accuracy from three strategies (IDTS, DTS, STS) is depicted. IDTS yields 68.3\%, improving by 3.06\% and 1.43\% compared to STS and DTS. IDTS's edge lies in dynamic threshold adjustment for each class, ensuring adequate pseudo label generation for every category.

\subsubsection{Comparisons of Different Prediction Strategies}
\begin{figure}[t!]
  \centering
    \begin{subfigure}[t]{0.48\columnwidth}
    \includegraphics[width=\textwidth]{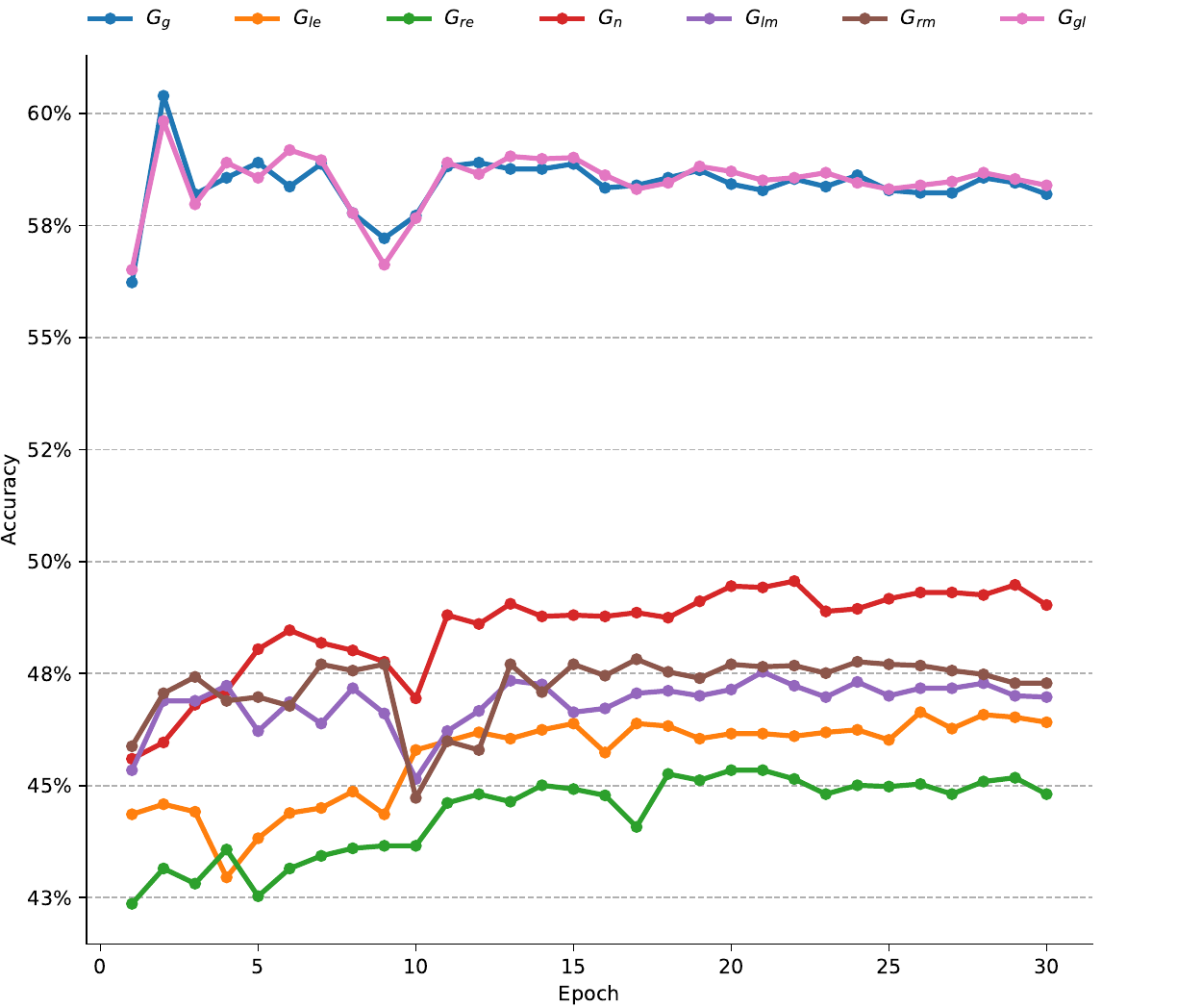}
    \caption{}
    \label{fig:training-curves}
    \end{subfigure}
    \hfill
    \begin{subfigure}[t]{0.48\columnwidth}
    \includegraphics[width=\textwidth]{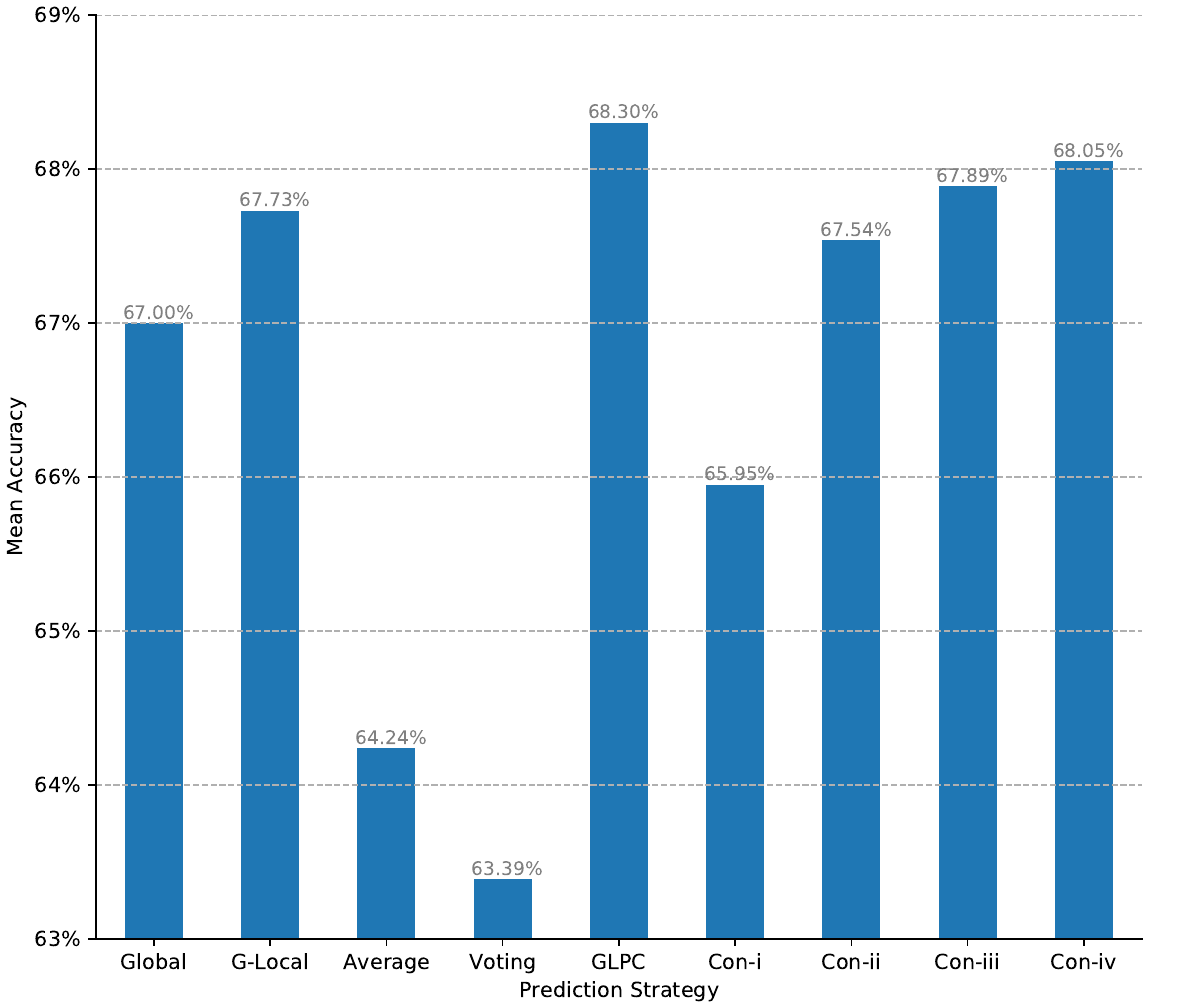}
    \caption{}
    \label{fig:prediction-mean-acc}
    \end{subfigure}
    \caption{Ablation study of GLPC. (a) Accuracy curves of the seven classifiers during training. RAF-DB is used as the source domain, ResNet50 as the backbone, and FER2013 as the target domain. $G_{le}$ is short for the classifier whose input feature is the left eye, and the same goes for the other classifiers. (b) Mean accuracies of different prediction strategies on five target domain datasets.}
    \label{fig:Ablation-GLPC}
\end{figure}

\begin{table}
  \centering
    \caption{Accuracies of different prediction strategies. The best results are in bold.}
    \label{tab:Prediction-Strategy}
  \begin{tabular}{c|ccccc}
  \toprule
  \centering  Methods & CK+ & JAFFE & SFEW2.0 & FER2013 & ExpW \\
  \hline
  \centering  Global & 86.05 & 60.56 & 55.28 & 60.34 & 72.78\\
  \centering  G-Local & 86.82 & 61.03 & 57.34 & 60.51 & 72.96\\
  \centering  Average & 80.62 & 59.15 & 55.73 & 58.57 & 74.25\\
  \centering  Voting & 80.62 & 59.62 & 52.52 & 57.07 & 73.34\\
  \hline
  \centering  GLPC & \textbf{87.60} & \textbf{61.97} & 58.26 & \textbf{60.68} & 73.00\\
  \hline
  \centering  Con-i & 79.84 & 59.15 & 57.80 & 58.59 & \textbf{74.40}\\
  \centering  Con-ii & 85.27 & 61.97 & 57.11 & 60.45 & 72.89\\
  \centering  Con-iii & 86.82 & 60.56 & \textbf{58.72} & 60.34 & 73.03\\
  \centering  Con-iv & 87.60 & 61.97 & 57.34 & 60.48 & 72.88\\
  \bottomrule
  \end{tabular}
  \vspace{4pt}
\end{table}

The AGLRLS framework incorporates a total of seven classifiers, as depicted in Fig. \ref{fig:Ablation-GLPC}(a) to showcase their classification performance. We introduce two single-classifier approaches (Global classifier $G_{g}$ and Global-Local classifier $G_{gl}$) and two multi-classifier prediction strategies (Average and Voting) for a comparative analysis against our proposed GLPC (Global-Local Prediction Consistency) strategy. The results, presented in Table \ref{tab:Prediction-Strategy}, demonstrate GLPC's superior performance across the datasets. Fig. \ref{fig:Ablation-GLPC}(b) visually confirms that GLPC outperforms the aforementioned prediction strategies in terms of Mean accuracy. This substantiates the effectiveness of the GLPC strategy in optimally balancing weights of diverse classifiers' prediction, thereby better fusing global and local facial information for more accurate expression prediction.

The order of classifier predictions is crucial in the GLPC strategy. Therefore, to validate its effectiveness, we establish four variations: Con-i, Con-ii, Con-iii, and Con-iv. Con-i fuses seven classifier scores and thresholds for prediction. Con-ii and Con-iii use global and combined global-local feature classifiers. If unsatisfactory predictions occur, they follow Con-i. Con-iv does the same sequentially, using global and combined classifiers, with a fallback to Con-i if needed. Con-iv closely resembles the GLPC strategy, differing in that GLPC sequentially employs global-local and global classifiers, while Con-iv follows the reverse order. As shown in Table \ref{tab:Prediction-Strategy}, GLPC outperforms the other four variants on CK+, JAFFE, and FER2013 datasets. The excellent accuracy displayed in Fig. \ref{fig:Ablation-GLPC}(b) further confirms the effective prediction order of GLPC in balancing classifier performance.

\section{CONCLUSIONS}
This paper has proposed a novel adaptive global-local representation learning and selection framework that aims to learn robust domain-invariant features and select the optimal fused prediction to mitigate the domain shift in cross-domain facial expression recognition. The proposed model incorporates the global-local adversarial adaptation and semantic-aware pseudo label generation to facilitate the learning of domain-invariant feature representation in an end-to-end manner. Additionally, it learns the optimal prediction label through a global-local prediction consistency learning strategy during the inference phase. The experimental results demonstrate that it is capable of largely increasing the accuracy of the CD-FER task. The experiments using different modules, pseudo generation strategies and prediction strategies, have validated the benefit of our proposed components to CD-FER.




\bibliographystyle{IEEEtran}
\bibliography{reference}
\vspace{-10 mm}
\begin{IEEEbiography}[{\includegraphics[width=1in,clip]{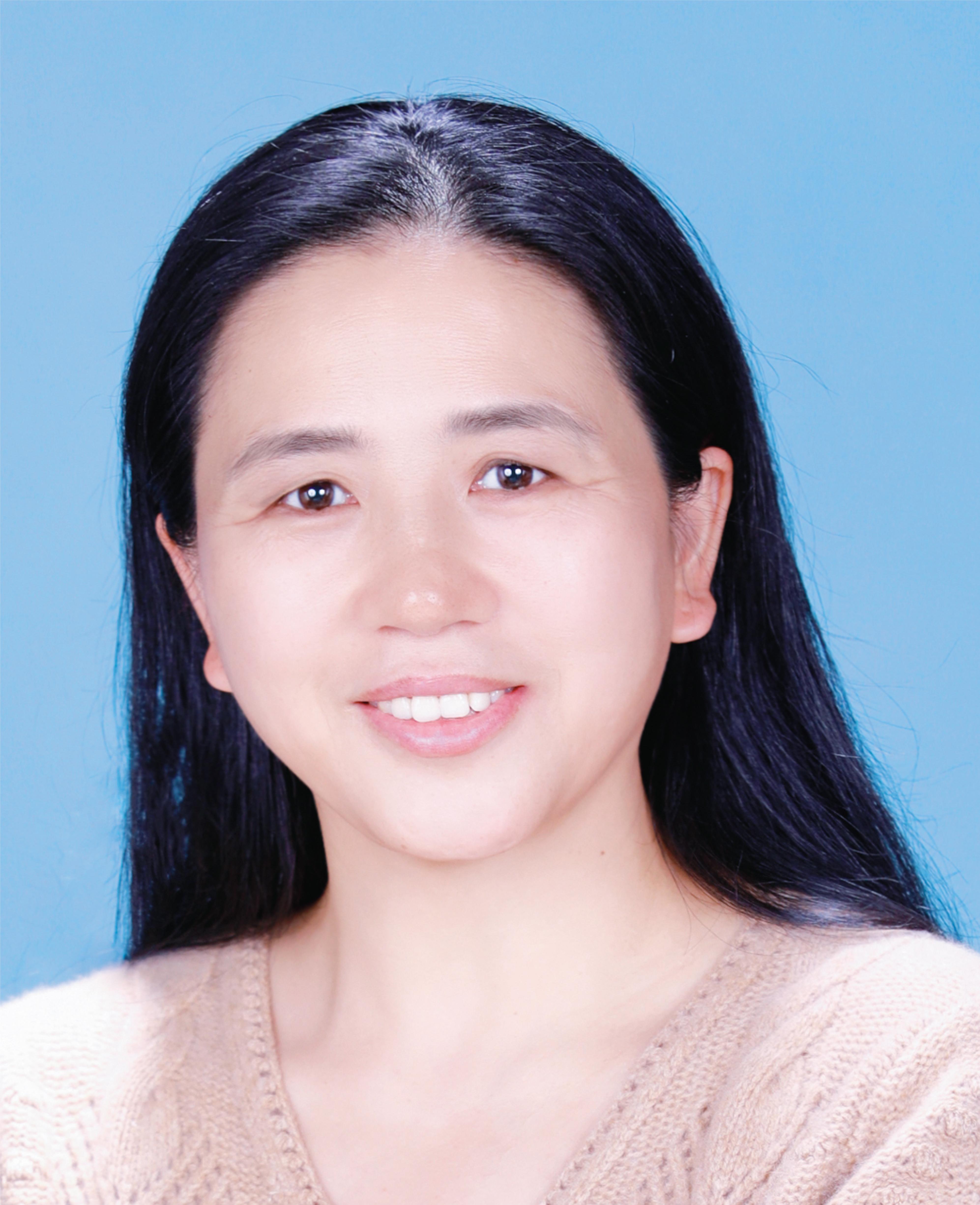}}]{Yuefang Gao} received the Ph.D. degree in computer science from South China University of Technology in 2009. She is an associate professor in College of Mathematics and Informatics, South China Agricultural University, China. Her research interests include computer vision, machine learning and data mining. She has authored and co-authored more than 20 papers in top-tier academic journals and conferences, including T-CSVT, PR, TKDE, ACM MM, etc. She has served as a reviewer for some academic journals and conferences. 
\vspace{-10 mm}
\end{IEEEbiography}

\begin{IEEEbiography}[{\includegraphics[width=1in,height=1.25in,clip,keepaspectratio]{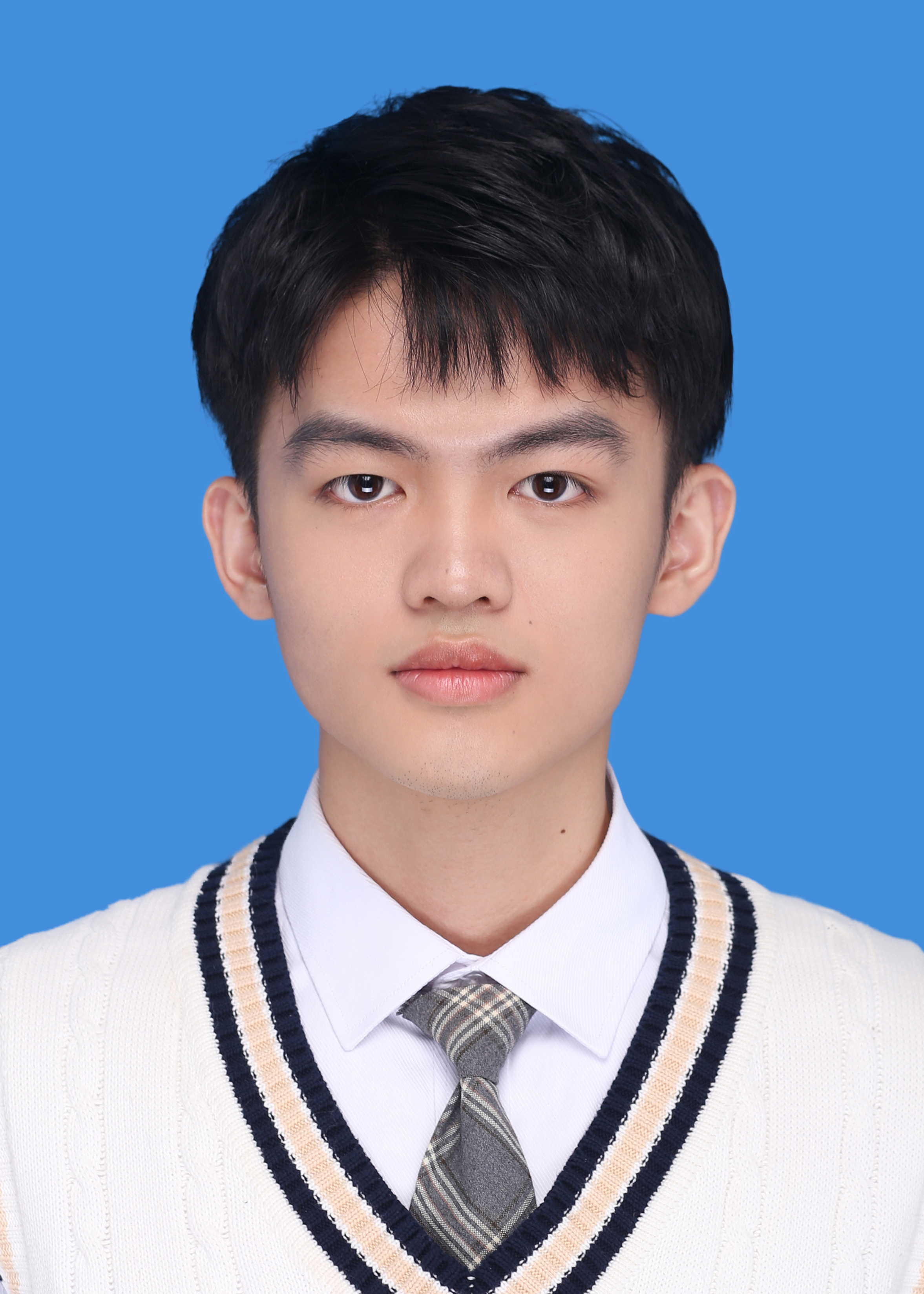}}]
{Yuhao Xie} received his bachelor's degree from South China Agricultural University, China in 2023. He is currently pursuing a master's degree at the Guangzhou Institute of Technology, Xidian University. His research interests include computer vision and deep learning. 
\vspace{-10 mm}
\end{IEEEbiography}

\begin{IEEEbiography}[{\includegraphics[width=1in,height=1.25in,clip,keepaspectratio]{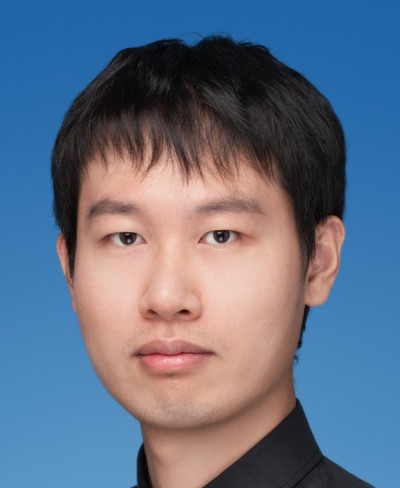}}]{Zeke Zexi Hu} is currently a Ph.D. candidate at the School of Computer Science, University of Sydney, Australia. He received his bachelor's degree from South China Agricultural University, China in 2014 and his M.Phil. degree from the University of Sydney in 2020. His research focuses on computer vision and deep learning. 
\vspace{-10 mm}
\end{IEEEbiography}

\begin{IEEEbiography}[{\includegraphics[width=1in,height=1.25in,clip,keepaspectratio]{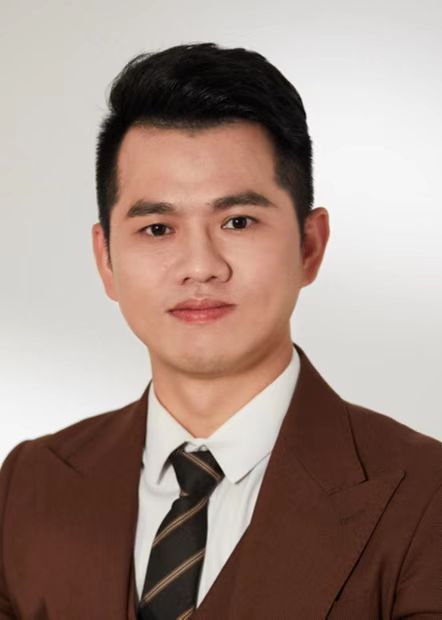}}]{Tianshui Chen} received a Ph.D. degree in computer science at the School of Data and Computer Science Sun Yat-sen University, Guangzhou, China, in 2018. Prior to earning his Ph.D., he received a B.E. degree from the School of Information and Science Technology in 2013. He is currently an associated professor in the Guangdong University of Technology. His current research interests include computer vision and machine learning. He has authored and coauthored approximately 40 papers published in top-tier academic journals and conferences, including T-PAMI, T-NNLS, T-IP, T-MM, CVPR, ICCV, AAAI, IJCAI, ACM MM, etc. He has served as a reviewer for numerous academic journals and conferences. He was the recipient of the Best Paper Diamond Award at IEEE ICME 2017. 
\vspace{-10 mm}
\end{IEEEbiography}

\begin{IEEEbiography}[{\includegraphics[width=1in,clip]{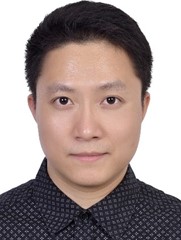}}]{Liang Lin} (Fellow, IEEE) is a full professor at Sun Yat-sen University. From 2008 to 2010, he was a postdoctoral fellow at the University of California, Los Angeles. From 2016--2018, he led the SenseTime R\&D teams to develop cutting-edge and deliverable solutions for computer vision, data analysis and mining, and intelligent robotic systems. He has authored and coauthored more than 100 papers in top-tier academic journals and conferences (e.g., 15 papers in TPAMI and IJCV and 60+ papers in CVPR, ICCV, NIPS, and IJCAI). He has served as an associate editor of IEEE Trans. Human-Machine Systems, The Visual Computer, and Neurocomputing and as an area/session chair for numerous conferences, such as CVPR, ICME, ACCV, and ICMR. He was the recipient of the Annual Best Paper Award by Pattern Recognition (Elsevier) in 2018, the Best Paper Diamond Award at IEEE ICME 2017, the Best Paper Runner-Up Award at ACM NPAR 2010, Google Faculty Award in 2012, the Best Student Paper Award at IEEE ICME 2014, and the Hong Kong Scholars Award in 2014. He is a Fellow of IEEE, IAPR, and IET.
 
\end{IEEEbiography}

\end{document}